\documentclass[letterpaper, 10 pt, conference]{ieeeconf}  

\IEEEoverridecommandlockouts                              
\usepackage{xcolor}

\usepackage{graphicx}
\usepackage{verbatim}
\usepackage{amsmath}
\usepackage{amssymb}
\usepackage{threeparttable}
\usepackage{booktabs}
\usepackage{multirow}
\usepackage{makecell}
\usepackage{subcaption}
\usepackage{comment}
\usepackage{booktabs}
\usepackage[longend,ruled,vlined, linesnumbered]{algorithm2e}

\SetCommentSty{mycommfont}
\SetKwInput{KwInput}{Input}                
\SetKwInput{KwOutput}{Output}              

\makeatletter

\def\algbackskip{\hskip-\ALG@thistlm}
\makeatother
\usepackage{pifont}
\usepackage{xcolor}
\usepackage{tabularx}

\DeclareMathOperator*{\argmaxA}{arg\,max}

\usepackage{threeparttable}
\usepackage{booktabs}
\usepackage{multirow}
\usepackage[
  separate-uncertainty = true,
  multi-part-units = repeat
]{siunitx}
\usepackage{tikz}

\usepackage{array}

\newcolumntype{B}{!{\hspace{-1ex}}c}
\newcolumntype{D}{!{\hspace{-2ex}}c}
\newcolumntype{A}{!{\hspace{-1ex}}l}
\usepackage{float}
\usepackage[noadjust]{cite} 

\makeatother

\newcommand{\mycomment}[1]{}

\title{\LARGE \bf
Find a Way Forward: a Language-Guided Semantic Map Navigator
}

\author{Authors}
\author{Zehao Wang*$^{1}$, Mingxiao Li*$^{2}$, Minye Wu$^{1}$, Marie-Francine Moens$^{2}$, Tinne Tuytelaars$^{1}$
\thanks{* Equal contribution}
\thanks{$^{1}$Zehao Wang, Minye Wu and Tinne Tuytelaars are members of the Electrical Engineering Department (ESAT-PSI) of
        KU Leuven, 3000 Leuven, Belgium
        {\tt\small zehao.wang@esat.kuleuven.be}}%
\thanks{$^{2}$Mingxiao Li and Marie-Francine Moens are members of the Computer Science Department of KU Leuven,
        3000 Leuven, Belgium
        {\tt\small mingxiao.li@kuleuven.be}}%
}

\begin{document}

\maketitle

\thispagestyle{empty}
\pagestyle{empty}


\begin{abstract}
In this paper, we introduce the {\em map-language navigation} task where an agent executes natural language instructions and moves to the target position based only on a given 3D semantic map. 
%
To tackle the task, we design the instruction-aware Path Proposal and Discrimination model (iPPD). 
Our approach leverages map information to provide {\em instruction-aware path proposals}, i.e., it selects all potential instruction-aligned candidate paths to reduce the solution space. 
Next, to represent the map observations along a path for a better modality alignment, a novel {\em Path Feature Encoding} scheme tailored for semantic maps is proposed. 
%
%
An attention-based {\em  Language Driven Discriminator} is designed to evaluate path candidates and determine the best path as the final result. Our method can naturally avoid error accumulation compared with single-step greedy decision methods.
%
%
%
%
Comparing to a single-step imitation learning approach, iPPD has performance gains above 
17\% 
on navigation success and 0.18 on path matching measurement nDTW in challenging unseen environments. 

\end{abstract}


\section{INTRODUCTION}
The general-purpose robot assistant of the future assists humans with daily tasks to reduce labor overhead, e.g.,~as a housekeeping or indoor service robot. An essential part of the human-robot interaction involves {\em language guided navigation}, i.e.,~enabling the robot to execute instructions given by a human to reach a target location. This requires the robot to interpret the 
natural language instruction, ground them in (usually visual) observations, and move accordingly. 
This is a challenging task that previously has been addressed mostly using single-step greedy decision methods, such as imitation learning or reinforcement learning (e.g.,~\cite{vlnce,krantz2021waypoint}) combined with pre-trained networks for encoding the visual~\cite{ddppo,deng2009imagenet} and language inputs~\cite{devlin2018bert,li2019robust}. 
These approaches are fully learning-based, i.e., mapping inputs to outputs without performing any geometric reasoning or relying on other types of domain knowledge. They are also 'greedy' in the sense that decisions are taken locally ("what is the next best move?") rather than planning ahead, overlooking the entire trajectory. This situation reduces error tolerance and does not allow considering alternative routes.

In contrast, robot navigation 
has been an active area of research for decades,
with well-established path planning methods on occupancy maps that have proven their value in a myriad of real-world applications. The question then naturally arises how this can be leveraged in language-guided navigation.

Some recent works for language-guided navigation include a pre-exploration phase (e.g.,~\cite{reinforced}), during which the robot explores the environment beforehand. This is introduced to tackle the domain gap by 
finetuning the visual encoder 
and has been shown to have a significant impact on the performance. Such a pre-exploration phase seems reasonable for most practical applications. 
%
\begin{figure}[t]
\centering
    \includegraphics[width=1.0\linewidth]{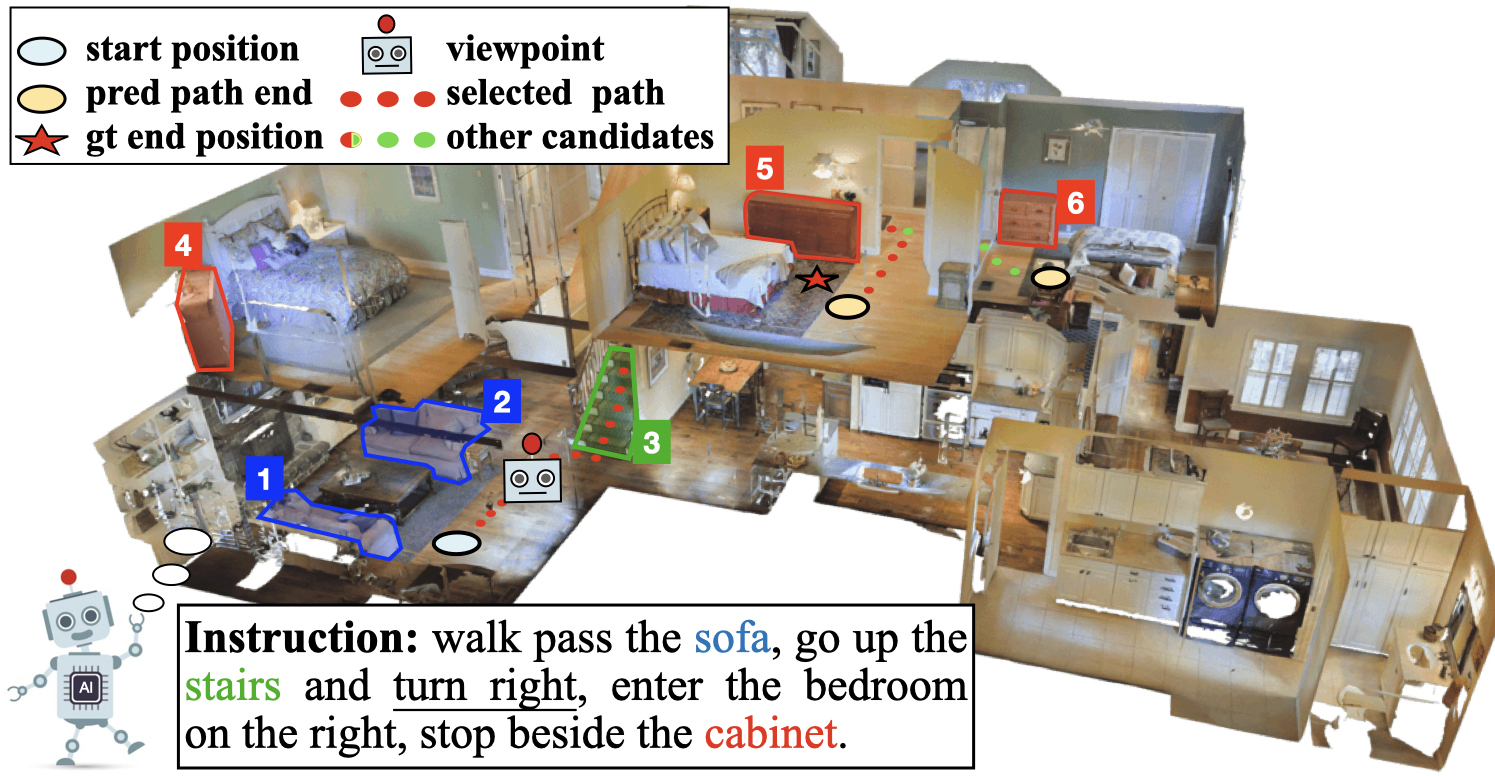}
\caption{In an instruction-guided navigation task, our agent first extracts key components (objects and turning actions) from the instruction, then proposes paths on the 3D semantic map. Since there can be multiple candidates, the agent encodes viewpoints along each path and ranks the paths to obtain the final prediction. One viewpoint feature for this example contains key components 1,2,3 and other objects surrounding it. Paths that pass by key components 5 and 6 are both valid, but the former one suits the instruction best.}
\label{fig:Teaser}
\end{figure}
Besides, in most scenarios, the layouts of the environment are mostly stationary. We usually need to carry out multiple navigation tasks in the same scene. For example, we may give orders to a housekeeping robot who always works in an indoor space, like at your home; Or the service robot of a hotel needs to deliver things to customers inside the building. In such scenarios, we can pre-scan the environments, obtain detailed semantic navigation maps, and rely on these for the actual navigation -- in combination with obstacle avoidance to cope with dynamic objects. These maps may include not only the geometry of the environment, as in a traditional occupancy map, but also its semantics, such as the location of the main objects, which are helpful to ground the language instructions. 

%
%
In this paper, we introduce the \textbf{Map-Language Navigation task (MLN)}, which carries out navigation based on the 3D semantic map of environments and natural language instructions. Compared with traditional visual-language navigation (VLN) using RGBD observation during navigation, we see two main benefits of conducting navigation based on a 3D semantic map. First, the map provides global information about the environment, which prevents the agent from being stuck in local decisions and makes it possible to plan the whole trajectory directly. Second, the maps of different environments share the same semantic definitions. This helps the agent to generalize to a previously unseen environment.   
To tackle this task, different from previous works~\cite{levit2007interpretation,vogel2010learning} which simply treated the map as additional visual inputs, we propose an \textbf{instruction-aware Path Proposal and Discrimination model (iPPD)} following planning and scoring 
pipeline, which is proved, in our experiments, to be a better way of leveraging map potential. 
Our model uses a modular design that consists of three main components. 
In the \textbf{First} module, we adopt deterministic algorithms to analyze the instructions and leverage them for path proposals. 
In particular, we extract key components (objects and turning actions) from the instruction and use them to reduce the solution space. For instance, in Fig.~\ref{fig:Teaser}
%
candidate path by key component \textit{object 4} is filtered out since it violates key component ``turn right".
%
\textbf{Second}, we introduce a semantic feature extraction module to encode the environment's information along a path into a feature sequence. 
%
iPPD perceives the environments in an egocentric perspective for a set of key points along a path candidate. 
A novel feature extraction scheme is introduced to obtain permutation invariant features at the object level. 
%
%
%
We also take the low-level features of the path, e.g., moving directions and locations, into account. 
\textbf{Third}, we use a multimodal transformer encoder to fuse path features and instructions and further score paths. 
The final prediction is determined according to the ranking of these scores.

We evaluate and compare our iPPD method with traditional single-step decision methods. Comprehensive experiments show that the proposed method can leverage the maps well and can produce effective results. 


\section{RELATED WORK}

%


\textbf{Vision-and-Language Navigation.} In the VLN~\cite{anderson2018vision} task, an agent needs to navigate to a goal location in a photo-realistic virtual environment following a given natural language instruction. 
Research on the  VLN task has made significant progress in the past few years. Attention mechanisms across different modalities are widely used to learn the alignment between vision and language, hence boosting the performance~\cite{vlnce,reinforced,huang2019multi,landi2021multimodal,ma2019self,ma2019regretful,landi2019embodied,hong2020sub}. The improvements of VLN also come from new learning approaches, such as imitation learning (IL)~\cite{wang2018look}, reinforcement learning (RL)~\cite{reinforced,ddppo}, or the ensemble of both IL and RL~\cite{krantz2021waypoint}.
Another line of work focuses on exploiting a map of the environment in navigation task~\cite{gupta2017unifying,seymour2021maast,chaplot2020learning,chen2019learning,tan2022self}. Previous works mainly follow the pipeline of building the 2D map using RGB and depth images on the fly first, then integrating the built map into the model as additional visual input. Different works construct maps of different quality, which makes it hard to measure whether the performance difference comes from the navigation algorithm or the map quality. Furthermore, the 2D map inevitably drops some important environment information; 
for example, when two objects are stacked at one location, only one will appear. 
In this work, we decouple the visual-language navigation task to map construction and map navigation. We focus on the second part and propose a language-guided map navigation task where a perfect 3D map of the environment is given and the agent can only use the language and map information, aiming to provide a base for future research to focus on the design of language-guided navigation algorithms.  

\textbf{Multi-modal Transformer} The transformer model has achieved great success on natural language processing and vision-language tasks. \cite{devlin2018bert} first pre-train the BERT model on large scale text data and achieve state of the art performances on a wide range of natural language processing (NLP) tasks. Inspired by the great success of BERT on NLP, a lot of researches extend the transformer model to process both vision and language information. \cite{devlin2018bert,tan2019lxmert} propose a two-stream BERT model to first encode texts and images separately then fuse the two modalities via a cross modal attention layer. Another line of work introduces single stream multimodal BERT, which directly processes both vision and language information simultaneously using cross modal attention layers~\cite{chen2020uniter,li2020unicoder,li2019visualbert}. Different from previous works which mainly focus on studying transformer architectures that handle vision and language data, in this work we further exploit the transformer model and extend it to process language and 3D map information. 

\section{OVERVIEW}
Given 
the start position $v_0$ in an environment, a natural language instruction $T$ and the voxelized 3D semantic map $\mathbf{M}$ of the environment, our model (iPPD) aims to find a path  
$\hat{\boldsymbol{m}}$ following the instruction.
%
We formulate it as:
\begin{equation}
\begin{aligned}[c]
    \hat{\boldsymbol{m}} = F_{\text{iPPD}}(\mathbf{M}, T, v_0), 
\end{aligned}
\end{equation}
where $F_{\text{iPPD}}(\cdot)$ is the proposed navigator. Each voxel in $\mathbf{M}$  has three labels, i.e., navigable (binary label) and object label (from the label set $\mathcal{Q}$).

\begin{figure*}[t]
\centering
    \includegraphics[width=0.90\linewidth]{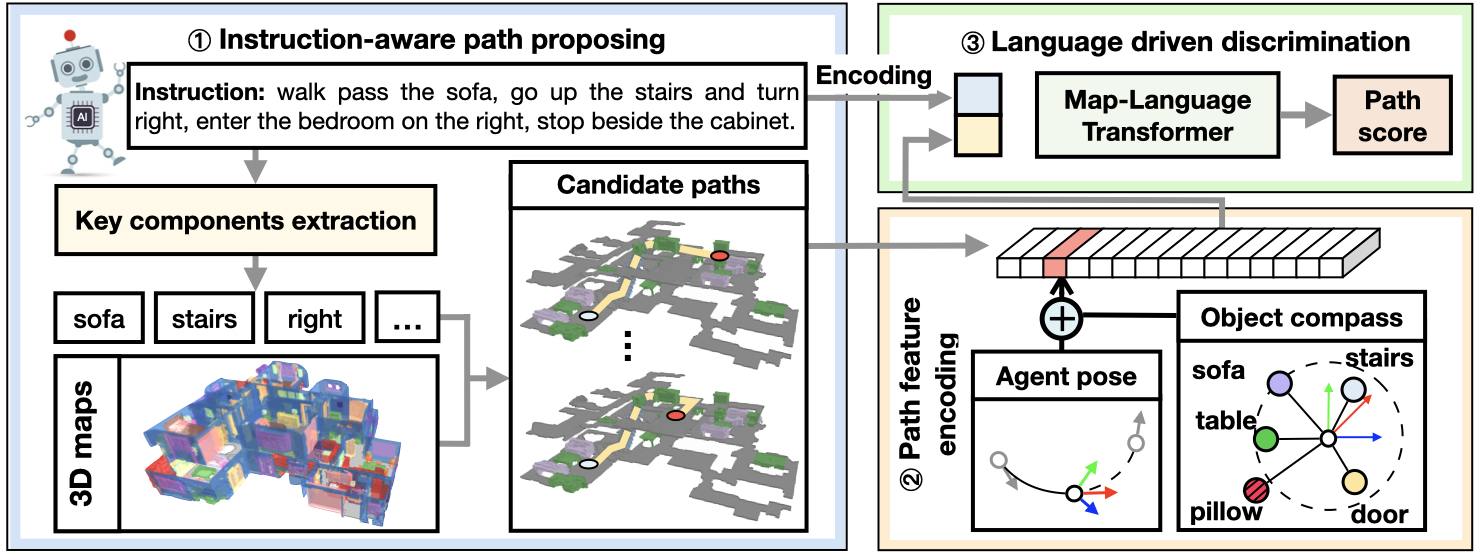}
\caption{\textbf{Data flow of instruction-aware path proposals discriminator (iPPD)}. Natural language instructions are leveraged to extract key components for path proposing. The paths are encoded by intermediate agent poses and surrounding objects. A multimodal transformer model is trained to score paths and select the highest one as the final prediction.}
\label{fig: pipeline}
\vspace{-2ex}
\end{figure*}

We divide the proposed pipeline into three modules, as demonstrated in Figure~\ref{fig: pipeline}. 
%
%
Specifically, in the first part (\S~\ref{sec:proposal}), we design an instruction-aware path proposing method to obtain candidate paths. 
%
%
In the second part (\S~\ref{sec:path_feature}), 
we discretize each candidate path and embed local observations into a feature sequence. 
iPPD extracts feature from egocentric views of the local 3D semantic map $\mathbf{M}$ along the path. Together with low-level features, such as moving directions and locations, we construct the overall path features.
%
%
Lastly, based on these path features, a language-driven discriminator (\S~\ref{sec:discriminator}) scores each candidate path, and the path with the highest score is taken as the final answer.

\section{METHOD}
Exploiting a semantic map for language-guided navigation is nontrivial. Naively integrating maps into existing imitation learning approaches~\cite{vlnce} does not have satisfactory performance, as demonstrated in experiments (\S~\ref{sec:exp_main}). Therefore, we design a novel way to handle map-language navigation by path proposing and scoring. 

\subsection{Instruction-aware Candidate Path Proposing}\label{sec:proposal}
In language-guided navigation, the agent has to navigate to the destination following the given instruction. Landmarks and moving actions mentioned in the instruction provide clues for finding the correct trajectory.  
We first propose an algorithm to extract key components. Then we do optimal path searching constrained by the extracted key component sequence with the A* algorithm to obtain path proposals.

\noindent\textbf{Key component extraction.} 
The key component sequence 
is constructed from two aspects, i.e., object landmarks and moving actions. 
Since similar meaning can have various forms of expression in natural language, our key component extraction 
relies on part-of-speech (POS) tags~\cite{collins2002pos,koo2008browncluster}, such as ``NNS" stands for plural nouns, 
and synsets\footnote{Synset represents the actual sense of a word in a sentence, which is the core component of WordNet. Different words can share the same synset.} of word tokens in each instruction. 
We process each instruction by Algorithm~\ref{code: nlp}. 
%
A manually verified reference dictionary $\mathbf{S}_{obj}$ mapping the object label $q$ to its synset is prepared beforehand.
In the landmark extraction step, we identify the synset $s$ of each noun $w.text$ in the context of instruction $T$ with the technique of Word Sense Disambiguation (WSD)~\cite{Lesk1986wsd}. Then we iteratively assign the closest object label $\tilde{q}$ to each noun measured by their synset-level WuPalmer similarity score\footnote{WuPalmer similarity measures how close two synsets are in the WordNet tree, it can handle hypernym/hyponym relations between two synsets.}. 
Noun words with scores higher than a threshold $\gamma$ are treated as key components. To keep a balance between recall and precision, we manually tune $\gamma$ and set it to 0.85.
On the other hand, the moving action is recognized by pattern matching. 
We formulate it as a rule-based flow in Algorithm~\ref{code: nlp}, line 9-17. We consider ”turn left“ and ”turn right“ as important moving actions to constrain proposal paths. Turn left or turn right actions are identified if a sequence of tokens meet the following conditions: the first token is a verb, a 'left' or 'right' token can be found within the next $k=3$ tokens and the immediate following token of 'left' or 'right' is neither a noun nor a 'turn' token.

\noindent\textbf{Instruction-aware path proposals} 
%
After extracting the temporally ordered key-component sequence from the instruction, we then propose candidate paths. We apply A* algorithm to plan a candidate path from a start position to a randomly sampled end position nearby the last object landmark and constrain the path to visit the key components in the extracted sequence in the temporal order. Since each object category might have multiple instances in the semantic map, this results in an average of 560 candidate paths; the worst case even has more than 80,000 paths.
We set three constraints to prune the number of paths. 
(1) Candidates violating extracted moving actions are removed. For instance, when we process the moving action ``turn left" with a following object landmark ``chair", we only keep the positions of chairs to the left of current agent orientation as the next visited points. (2) We set an upper bound on the search range towards the next landmark by the distance of five meters. We assume that landmarks that appear further than five meters are hard to follow and do not usually appear in an indoor navigation scenario. (3) We group nearby objects of the same label by density-based spatial clustering algorithm DBSCAN~\cite{Ester1996DBSCAN}. The landmark positions are replaced by their cluster core samples. We set the maximum distance between two samples of DBSCAN to 0.5 meters (a larger distance increases the risk of merging objects across rooms). This further merges similar paths to reduce the candidate number. The overall reduction rate is 85\%.
In case the model fails to extract valid key components from the instruction, we use random paths planned by A* algorithm with the same start position but different endpoints within twenty meters (maximum distance in indoor navigation dataset) as candidates. 
The full candidate set is denoted by $\mathcal{P} = \{\boldsymbol{m}_{v_0,e} \mid e \in \mathcal{V}_e \}$.

\subsection{Path Feature Encoding}\label{sec:path_feature}
Next, we propose a path feature embedding scheme and apply it to each candidate path in the semantic map $\mathbf{M}$. Ideally, the path feature $\mathbf{\Gamma}$ represents the environment context along the path and will be aligned with the language representation to verify if this path matches the given instruction. iPPD discretizes each path into multiple key points 
and encodes the local context of each key point along the path to
form a feature sequence in temporal order. Specifically, for each keypoint feature, we introduce 
an object compass to perceive the local environment as illustrated in Figure~\ref{fig: pipeline}. 
%
%

The object compass $\mathcal{C}_{\text{obj}} = \{(x_i, y_i, z_i, \mathbf{w}_i)\}$ contains objects near the agent within a $3$ meters radius. We record object position $(x_i, y_i, z_i)$ in the egocentric coordinate system centered at the agent. The object compass is treated as a 3D point cloud in a 3D egocentric view. For each point in the cloud, we embed the object's class with a 50-dimension GLOVE word embedding~\cite{pennington2014glove} $\mathbf{w}_i$. 
To obtain permutation invariant representation towards points order, we use PointNet~\cite{qi2017pointnet} to encode the object compass: 
\begin{equation}
     \mathbf{h}= \text{PointNet}(\mathcal{C}_{\text{obj}}), 
\end{equation}
where $\mathbf{h}\in \mathbb{R}^{d}$ is the $d$-dimensional feature of the local object's context of a keypoint.
%

We also incorporate low-level information, i.e., agent pose, into the feature sequence. An agent pose feature $\mathbf{s}$ contains location $\mathbf{x}$ and orientation $\theta$ at a key point:
\begin{equation}
    \mathbf{s} = \text{FFN}_{ap}([PE(\mathbf{x}), PE(\theta)]), 
\end{equation}
where $[\cdot]$ represents concatenation, $PE(\cdot)$ is the agent pose positional encoding function~\cite{mildenhall2020nerf}, and $\text{FFN}_{ap}$ is a single-layer feed-forward network projecting agent position into the $d$-dimensional feature space.
By gathering features of all key points, we obtain the path feature $\mathbf{\Gamma}=\{ (\mathbf{h}_i, \mathbf{s}_{i}) \}$. 
\begin{algorithm}[t]
\DontPrintSemicolon
\SetNoFillComment
\SetAlCapHSkip{0pt}
\KwInput{\small $T$, $\gamma$, $k$ \tcp*[h]{text, score thr, phrase limit}}
\KwOutput{$Y$ \ \ \ \tcp*[h]{key component sequence}}
\KwData{$\mathbf{S}_{obj}$ \ \ \ \tcp*[h]{mapper from object to synset } } 
$\hat T \gets \text{POSTagging}(T)$ \; 
\textbf{\upshape Set} $Y$ \textbf{\upshape to} empty list \;
\ForEach{w \textnormal{\bf in} $\hat T$}{
     \tcp*[h]{Landmark extraction step } \;
     \uIf{w.tag \textnormal{\bf is in} \textnormal{[NN, NNS]} }{
        $s = \text{WSD}(\textit{w}.text, T)$ \;
        $\tilde{q} = \argmaxA_{q \in \mathcal{Q}} \text{WuPalmer}(s, \mathbf{S}_{obj}(q))$ \;
        \If{$\text{\normalfont WuPalmer}(s, \mathbf{S}_{obj}(\tilde{q})) > \gamma$}{
          $Y.append(\tilde{q})$ \;
        }
     }
     \tcp*[h]{Moving action recognition step } \;
     \ElseIf{w.tag \textnormal{\bf is in} \textnormal{[VB, VBZ, VBP]} }{
        $i \gets \textit{w}.index$\;
            \ForEach{j \textnormal{\bf in} \textnormal{[$i+1, i+ k$]}}{
            \If{$\hat T[j]$ in \textnormal{[`left', `right']}}{
                \uIf{$\hat{T}[j+1]$.tag \textnormal{\bf not in} \textnormal{[NN, NNS]}}{
                    $Y.append(\hat T[j])$ \;
                }
                \ElseIf{$\hat{T}[j+1]$.text \textnormal{\bf is} `turn'}{
                    $Y.append(\hat T[j])$ \;
                }
                \textbf{Exit the loop}
              } 
            }
     }
    }
\caption{Key component extraction}
\label{code: nlp}
\end{algorithm}
\subsection{Language-driven Discriminator} \label{sec:discriminator}
We deploy a transformer-based model~\cite{vaswani2017attention} as \textit{Language-driven Discriminator} $F_{tf}$ for scoring paths. 
The natural language instruction $T$ is tokenized and embedded by GLOVE~\cite{pennington2014glove} embeddings $\psi(\cdot)$. 
We project the word embeddings to feature vectors using a single-layer feed-forward network $\text{FNN}_{lang}(\cdot)$.
\begin{equation}
     \mathcal{T} = \text{FNN}_{lang}(\psi(T)), 
\end{equation}
where $\mathcal{T}$ is the representation of word sequence.

\begin{figure}[t]
\centering
      \includegraphics[width=1.0\linewidth]{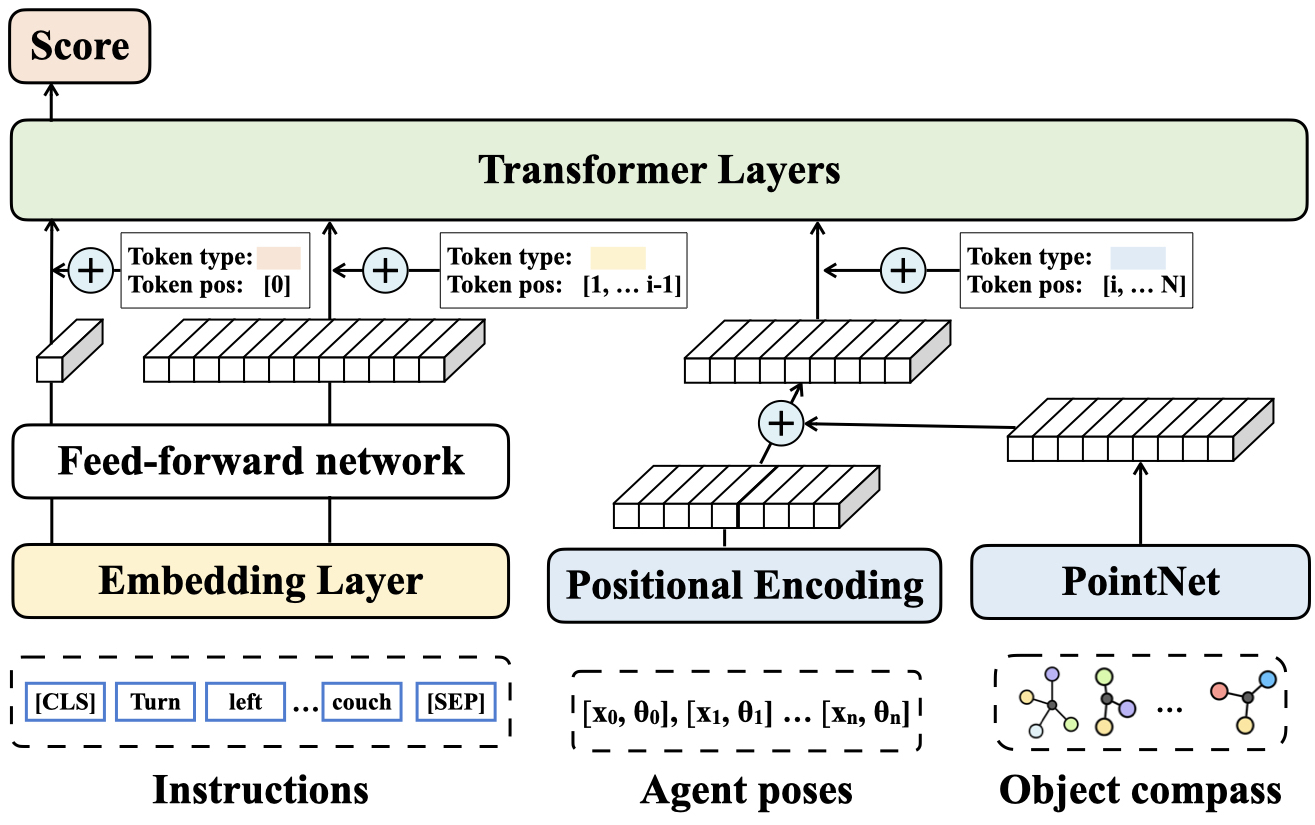}
\caption{\textbf{Language-driven discriminator architecture.} The model takes the concatenation of the $[CLS]$ special token, instruction token sequence and object point features as input and is trained using both score prediction MSE loss and a masked language model loss.}
\label{fig:network}
\end{figure}

The $F_{tf}$ has $N_t$ encoder layers as shown in Figure~\ref{fig:network}. 
It takes a concatenated sequence of $\mathcal{T}$ and $\mathbf{\Gamma}$ as inputs and predicts a score for the corresponding path. 
A learnable special token embedding $[CLS]$ is put at the front of the input sequence to aggregate context information. The \textit{Language-driven Discriminator} is formulated as
\begin{equation}
\begin{aligned}[c]
     \hat y = F_{tf}( [\text{[CLS]}, \mathcal{T}, \mathbf{\Gamma}] )
\end{aligned}
\end{equation}
where $\hat y$ is the prediction score projected from the $[CLS]$ token embedding. 
We pick the highest score path as the predicted path $\hat{\boldsymbol{m}}$ in inference.

During training, the ground truth score of each candidate path is set to be the combination of normalized Dynamic-Time Warping (nDTW)~\cite{ndtw} and Euclidean distance between candidate endpoint $v_e$ and golden target point $v_{gt}$. 
%
Specifically, nDTW, as a path shape similarity metric, indicates how well the candidate path matches the ground truth path. 
The target score $y$ is the linear combination of them, formulated as:
\begin{equation}
    y= \lambda \cdot \text{nDTW}(\hat{\boldsymbol{m}}_{s,e}, \boldsymbol m_{gt}) + (1-\lambda) \cdot \text{D}(v_e, v_{gt})
\end{equation}
where $\lambda$ is a coefficient to balance these two components and $\boldsymbol m_{gt}$ is the ground truth trajectory annotated in the dataset. 
In addition, we apply the Masked Language Prediction task ~\cite{devlin2018bert} to enhance the model's context-awareness. 
We randomly mask 15\% of the tokens in the instruction and train the model to predict the masked tokens during training. 
%
The total training loss $\mathcal{L}_{\text{total}}$ is formulated as:
\begin{equation}
    \mathcal{L}_{\text{total}}= \mathcal{L}_{MLM} + \mathcal{L}_{MSE}, 
\end{equation}
where $\mathcal{L}_{MLM}$ is the Masked Language Prediction loss, and $\mathcal{L}_{MSE}$ is the mean square error between $y$ and $\hat y$.

\section{Experiments}
\subsection{Dataset}
We extend the existing VLNCE~\cite{vlnce} dataset for this map-based language-guided navigation task and augment the dataset with our constructed 3D semantic maps but drop the RGBD visual input. We follow the original dataset splits 
and rename the splits as train, evaluation seen, and evaluation unseen, respectively. 
The environments in the evaluation seen set have been observed in the training phase, but the instructions are not, while both environment and instruction are not observed during training for episodes in the evaluation unseen set. 

\subsection{Implementation Details}

\noindent\textbf{Semantic Map Generation.}
Since the instruction annotations 
are from VLNCE~\cite{vlnce}(version 1\_2), we construct semantic maps on its corresponding ninety Matterport environments~\cite{Matterport3D}. 
We first interpolate the semantic mesh of Matterport environments to construct fine-grained semantic point clouds. The 3D semantic maps $\mathbf{M}$ are constructed by voxelizing semantic point clouds with a resolution of 0.05 meters. The semantic label of each voxel is the most frequent instance id and instance category 
among incoming points. 
Each instance category $q$ is one of the forty object categories $\mathcal{Q}$. The navigable areas are constructed from NavMesh of Matterport~\cite{Matterport3D}. We add a binary value for each voxel, indicating whether it is navigable or not during path planning.  

\noindent\textbf{Model setup.} Our method has three modules, the path proposal generator, the path feature encoder, and the path discriminator. The the path proposal generator is a fully CPU-intensive module that, in parallel, extracts candidate paths and point-wise object compasses. Path features are encoded by Feed-forward networks and PointNet as aforementioned. The results are batched and sent to a 12-head 6-layer transformer model for scoring. 
We optimize our model using AdamW~\cite{loshchilov2018adamw} optimizer with a learning rate of 1e-4 and train it on a single NVIDIA-P100 GPU for 12 hours.

\subsection{Evaluation Metrics}
We evaluate results
using the standard evaluation metrics in visual navigation and visual-language navigation tasks \cite{anderson2018evaluation,anderson2018vision,magalhaes2019effective}: trajectory length (TL), navigation error - average distance to goal in meters (NE), normalized dynamic-time warping (nDTW), oracle success (OS) which measures the percentage of predicted trajectories that pass the target point, success rate (SR) and success weighted by the normalized inverse of the path length (SPL). We choose nDTW and SR as our primary metrics, 
as they cover two important aspects of the navigation task: (a) predicted and ground truth path similarity, and
(b) accuracy of reaching the target location.

\begin{table*}[t]
\centering
    \begin{threeparttable}
        \begin{tabular}{A BBBBBB BBBBBB}
        \toprule
         \multirow{2}{*}{\textbf{Model}} & \multicolumn{6}{c}{\textbf{Evaluation\_Seen}} & \multicolumn{6}{c}{\textbf{Evaluation\_Unseen}}  \\
        \cmidrule(lr{1em}){2-7}  \cmidrule(lr{1em}){8-13} 
         & TL $\downarrow$ & NE $\downarrow$ & nDTW $\uparrow$ & OS $\uparrow$ & SR $\uparrow$ & SPL $\uparrow$ &  TL $\downarrow$ & NE $\downarrow$ & nDTW $\uparrow$ & OS $\uparrow$ & SR $\uparrow$ & SPL $\uparrow$  \\
        \midrule
        
        \textbf{Seq2Seq-MAP} 
         & 9.732 & 8.312 & 0.453 & 0.383 & 0.263 & 0.239 & 9.470 & 8.554 & 0.416 & 0.308 & 0.209 & 0.182 \\
         \textbf{CMA-MAP} 
         & 9.566 & 7.711 & 0.443 &  0.337 &  0.225&  0.197 & 9.355 & 8.195 & 0.412 & 0.284 & 0.184 & 0.158 \\
        
        
          \textbf{iPPD}  & 8.207 & 5.622 & 0.646 & 0.500 &  0.437 & 0.417 & 8.035 & 6.215 & 0.604 & 0.481 & 0.380  & 0.357 \\ 
         
        \bottomrule
        \end{tabular}
\end{threeparttable}
\caption{Performance in language guided navigation on 3D semantic maps. We implement three baselines, Sequence to Sequence generation model (Seq2Seq) and Cross Attention model (CMA) are imitation learning methods following common implementations~\cite{vlnce,krantz2021waypoint}, map language navigator (iPPD) is our newly proposed model. 
} 
\label{tab: main}
\end{table*}

\subsection{Main Results} \label{sec:exp_main}
We implement three baseline models to benchmark this new map-language navigation task. The first is the sequence-to-sequence model~\cite{vlnce,krantz2021waypoint} which is widely used in sequential decision-making tasks such as VLN. 
It jointly encodes map observation at each point and instruction into one feature vector and uses recurrent neural network GRU~\cite{cho-etal-2014gru} for action sequence prediction. Cross Attention Model (CMA) is another model that has proved successful in VLN task. The model conducts cross attention between language and observation at each time step to learn better multimodal alignment. We follow the previous work~\cite{vlnce} and train both sequence-to-sequence and cross-attention models using imitation learning. 
The last is our proposed iPPD model, which proposes potential instruction-aligned paths in 3D maps and uses transformer networks to do multimodal fusion and scoring. 
Table~\ref{tab: main} shows iPPD outperforms the other two baselines by a large margin. Specifically, in the challenging unseen environment, the iPPD surpasses the Seq2Seq and CMA models by $17 \%$ and $20 \%$ in SR. Our model can also conduct better instruction-aligned navigation, which is proved by the better performance in nDTW -- iPPD and it achieves around $0.19$ higher score than both Seq2Seq and CMA.

\subsection{Are both path proposal and discriminator helpful?}
This experiment studies the effectiveness of our instruction constrained path proposal algorithm and transformer-based path discriminator. To this end, we first compare the results of random path selection from the map with random path selection from the candidate path pool proposed by our path proposal algorithm. We named these two different algorithms 
%
random path and iPPD w/o path discriminator,
respectively. We then further compare the performance of our iPPD with the iPPD w/o path discriminator algorithm. The results are presented in Table.~\ref{tab: path discriminator}

In the first experiment, we conducted both random path and iPPD w/o path discriminator algorithms. For random path, we leverage the training dataset statistic that path lengths roughly follow a normal distribution with a mean of 8.89 meters and a standard deviation of 2.67. The lengths of randomly sampled paths follow this length distribution. As for the iPPD w/o path discriminator, we uniformly sample one path from the candidate path pool. A path is considered a success if it ends within three meters from the goal position. Table \ref{tab: path discriminator} illustrates that random path performs at least $15\%$ SR worse on both seen and unseen sets. Especially the path matching score nDTW drops significantly - more than 0.35 on both seen and unseen set. It indicates that our instruction-aware proposal strategy can effectively reduce the solution space, which further benefits the path discriminator.

In the second experiment, we compare the iPPD w/o path discriminator with our iPPD model. 
From Table~\ref{tab: path discriminator}, we could observe that although iPPD w/o path discriminator has achieved more than $20\%$ in SR in both seen and unseen environments, selecting the path with our proposed path discriminator can further boost the performance to $43.7\%$ and $38.0\%$ in SR in seen and unseen environments, respectively. The large improvements verify the effectiveness of our proposed discriminator. We further argue that our iPPD model can be seen as a two-step path planning strategy. The constrained path proposal plays the role of rough path selection, and the followed path discriminator conducts fine-grained path identification to select the best path.

\begin{table}[t]
 \centering
    \begin{threeparttable}
        \begin{tabular}{ l cc cc } 
        \toprule
         \multirow{2}{*}{\textbf{Model}} & \multicolumn{2}{c}{\textbf{Val-Seen} } & \multicolumn{2}{c}{\textbf{Val-Unseen} }\\
         \cmidrule(lr{1em}){2-3}  \cmidrule(lr{1em}){4-5} 
         & nDTW $\uparrow$ & SR $\uparrow$ & nDTW $\uparrow$ & SR $\uparrow$  \\
        \midrule
        \textbf{iPPD}   & 0.646  & 0.437  & 0.604  & 0.380  \\
        
         \textbf{{ w/o path discriminator}} &  0.397 &  0.219 &  0.413 &  0.203 \\
        
         \textbf{Random path}  & 0.025 & 0.039 & 0.030 &  0.040 \\

        \bottomrule
        \end{tabular}
    \end{threeparttable}

    \caption{Study of effectiveness of path proposal algorithm and  discriminator of iPPD. }
    \label{tab: path discriminator}
    \vspace{-2ex}
\end{table}

\subsection{How effective is each component in the discriminator?}

\begin{table}[t]
 \centering
    \begin{threeparttable}
        \begin{tabular}{ l cc cc } 
        \toprule
         \multirow{2}{*}{\textbf{Model}} & \multicolumn{2}{c}{\textbf{Val-Seen} } & \multicolumn{2}{c}{\textbf{Val-Unseen} }\\
         \cmidrule(lr{1em}){2-3}  \cmidrule(lr{1em}){4-5} 
         & nDTW $\uparrow$ & SR $\uparrow$ & nDTW $\uparrow$ & SR $\uparrow$  \\
        \midrule
        \textbf{iPPD}   & 0.646  & 0.437  & 0.604  & 0.380 \\
         
         \textbf{{ w/o agent pose}} & 0.594  & 0.415  &  0.590 &  0.364 \\
         \textbf{{ w/o object compasses}}  &  0.538 & 0.274 & 0.522  & 0.244   \\

        \bottomrule
        \end{tabular}
    \end{threeparttable}

    \caption{Ablation study for iPPD method on both the val\_seen and val\_unseen datasets. }
    \label{tab: Ablation study}
    \vspace{-2ex}
\end{table}

To verify the effectiveness of each component of our proposed path discriminator, we conduct ablation studies as shown in Table~\ref{tab: Ablation study}. Agent pose information regards the agent's state 
in the environment. This information is vital 
as we can see from the third line of Table~\ref{tab: Ablation study}: models trained without agent pose information have a drop of $2\%$ in SR on seen environments and $1.6\%$ on unseen environments. We next verify the effectiveness of encoding path feature using our proposed object compass.
The results in the last row of Table~\ref{tab: Ablation study} show the importance of point-wise context information. Removing the object compass from the model results in a significant drop in all evaluation metrics, the SR on seen environments and unseen environments decreases $16\%$ and $13\%$ respectively.
We summarize that both of the two proposed sub-modules are effective, especially semantic information encoded by object compass can significantly help the path discrimination process. 

\subsection{Generalization compared to visual-language setting}
\begin{table}[t]
\centering
    \begin{threeparttable}
        \begin{tabular}{AA B B B}
        \toprule
         & & Val\_Seen & Val\_Unseen & Diff \\
        \midrule
        
        RGBD & \textbf{Seq2Seq-DA*~\cite{vlnce}} & 0.338  &  0.252 & 0.086\\
        & \textbf{CMA-DA-PM*~\cite{vlnce}}  & 0.341 & 0.292 & 0.049\\
        & \textbf{HPN + DN*~\cite{krantz2021waypoint}}  & 0.482  & 0.363 & 0.119 \\
        
        \midrule
        
        Map & \textbf{Seq2Seq-DA-Map} & 0.263 &  0.209 & 0.054  \\
        & \textbf{CMA-DA-PM-Map} & 0.225 & 0.184 & 0.041\\
        
        & \textbf{iPPD}   & 0.437  & 0.380 & 0.057 \\ 
        
        \bottomrule
        \end{tabular}
    \begin{tablenotes}
    \item * stands for results reproduced by official checkpoints 
    \end{tablenotes}
\end{threeparttable}
\caption{
Navigation Success Rate (SR) and difference between seen and unseen validation set for various models.
} 
\label{tab: compare to vln}
\vspace{-2ex}
\end{table}
In visual-language navigation tasks, researchers argue that models developed between image observation and language are hard to generalize to unseen environments. As shown in Table \ref{tab: compare to vln}, the SOTA HPN model~\cite{krantz2021waypoint} trained with reinforcement learning on multiple GPUs for five days still suffers from a $10\%$ SR drop on unseen environments compared to seen set; however, this situation is less severe for our map-based models. We can observe that the difference is only $5\%$. At the same time, iPPD has superior performance in unseen environments compared to HPN. This indicates that the map has the potential to become a better modality to tolerate environment transfer, even for deep learning methods.

\section{Conclusion}
In this paper, we propose the map-language navigation task where an agent needs to navigate to an instruction-specified destination on a given 3D semantic map. The task is benchmarked with two commonly used language navigation models: Seq2Seq and CMA. The results of the two baseline models reveal the difficulties of the task. They also verify our assumption that the semantic map can benefit agents' generalization in unseen environments. We further introduce the iPPD model following a new proposal and discriminating pipeline to handle the task. Extensive experimental results show the superiority and effectiveness of our proposed rough and fine-grained two-step navigation algorithm. Furthermore,
our algorithm allows parallel training and inference, which could bring much convenience in real applications. We hope our work can inspire future research to focus on designing language-aware algorithms that can better leverage the 3D semantic map information.






\section*{ACKNOWLEDGMENT}
This project is supported by EWI Flanders AI Research program and KULeuven C1 project Macchina.

\bibliography{main.bib}
\bibliographystyle{IEEEtran}


\end{document}